\documentclass[lettersize,journal]{IEEEtran}
\usepackage{amsmath,amsfonts}
\usepackage{algorithmic}
\usepackage{algorithm}
\usepackage{array}
\usepackage[caption=false,font=normalsize,labelfont=sf,textfont=sf]{subfig}
\usepackage{textcomp}
\usepackage{stfloats}
\usepackage{url}
\usepackage{verbatim}
\usepackage{graphicx}
\usepackage{cite}
\usepackage{amsfonts, amsmath, amssymb, booktabs}
\usepackage{multirow}

\begin{document}

\title{Few-Shot Domain Adaptation for Charge Prediction \\on Unprofessional Descriptions}

\author{Jie Zhao,
        Ziyu Guan$^{*}$\thanks{* Corresponding author},
        Wei Zhao,
        Yue Jiang,
        Xiaofei He
\IEEEcompsocitemizethanks{
\IEEEcompsocthanksitem Jie Zhao, Ziyu Guan, Wei Zhao, and Yue Jiang are with the State Key Laboratory of Integrated Services Networks, School of Computer Science and Technology, Xidian University, Xi'an, Shaanxi 710071, China.
E-mail: \{jzhao1992@stu., zyguan@, ywzhao@mail., 22031212489@stu.\}xidian.edu.cn
\IEEEcompsocthanksitem Xiaofei He is with the State Key Lab of CAD\&CG, College of Computer Science, Zhejiang University, Hangzhou, Zhejiang 310058, China.
E-mail: xiaofeihe@cad.zju.edu.cn.}
}

\markboth{Journal of \LaTeX\ Class Files,~Vol.~xx, No.~xx, September~2023}%
{Zhao \MakeLowercase{\textit{et al.}}: Few-Shot Domain Adaptation for Charge Prediction on Unprofessional Descriptions}

\maketitle

\begin{abstract}
Recent works considering professional legal-linguistic style (PLLS) texts have shown promising results on the charge prediction task. However, unprofessional users also show an increasing demand on such a prediction service. There is a clear domain discrepancy between PLLS texts and non-PLLS texts expressed by those laypersons, which degrades the current SOTA models' performance on non-PLLS texts. A key challenge is the scarcity of non-PLLS data for most charge classes. This paper proposes a novel few-shot domain adaptation (FSDA) method named Disentangled Legal Content for Charge Prediction (DLCCP). Compared with existing FSDA works, which solely perform instance-level alignment without considering the negative impact of text style information existing in latent features, DLCCP (1) disentangles the content and style representations for better domain-invariant legal content learning with carefully designed optimization goals for content and style spaces and, (2) employs the constitutive elements knowledge of charges to extract and align element-level and instance-level content representations simultaneously. We contribute the first publicly available non-PLLS dataset named NCCP for developing layperson-friendly charge prediction models. Experiments on NCCP show the superiority of our methods over competitive baselines.
\end{abstract}


\section{Introduction}

\IEEEPARstart{S}{ignificant} research enthusiasm has arisen recently in utilizing artificial intelligence techniques to support legal decision-making tasks. Among these tasks, charge prediction, also referred to as crime prediction, aims to predict the ultimate charges adjudged by the court for suspects, relying on the factual details of legal cases. It can: serve as an intelligent assistant to offer readily accessible references for legal professionals to enhance their working efficiencies; act as a law study assistant for ordinary people who have zero or limited legal expertise \cite{surveyLegalAI}.

Typically, charge prediction is treated as a text categorization problem in the literature. Some works exclusively take fact descriptions of legal cases as input and concentrate on tackling the special textual characteristics (e.g., lengthier than regular texts) of legal texts \cite{rationale_augment,sattcaps,longtransformer}. Another promising direction is to explicitly incorporate legal knowledge to enhance the performance, such as designing charge-related features or attributes manually \cite{attributescharge,fewshot,qacharge}, incorporating law articles \cite{hierarchicalmatching,ladan}, exploiting constitutive elements of charges \cite{cecp,double_layer,inprovingljp}, and using textual content contained within the labels \cite{leapbank}. These works have demonstrated encouraging results in charge prediction. 

\begin{figure}[t]
	\centering
	\includegraphics[width=0.38\textwidth]{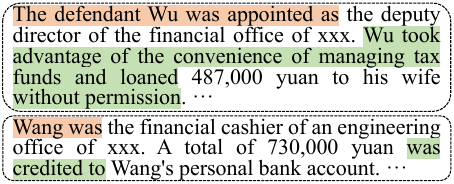}
	\caption{Fact descriptions (translated) with PLLS (top) and non-PLLS (bottom) for the charge ``embezzlement''. The red and green highlighted segments respectively represent textual distinctions between two domains when describing the occupation and conduct of the suspect.}
	\label{example}
\end{figure}

However, the above learned models can solely benefit legal professionals because all these studies only take into account fact descriptions with professional legal-linguistic style (PLLS). For lay people who are unfamiliar with the specialized characteristic of legal language, they can only present legal cases in a common style of language, which we call non-PLLS. A comparative example is shown in Fig.~\ref{example}. In the upper PLLS text, legal terms ``defendant" and ``appointed" are utilized when describing the suspect's occupation (highlighted in red), and ``took advantage of the convenience" and ``without permission" are employed when describing his conduct (highlighted in green). It is clear that the lower non-PLLS text do not contain such professional terms and expressions. This domain discrepancy due to style discrepancy between PLLS and non-PLLS will cause severe performance degradation for the previous SOTA models when they are applied to non-PLLS data. Luo \textit{et al.} \cite{fla} have demonstrated this phenomenon by applying the model trained on PLLS data to 100 legal cases with non-PLLS descriptions. However, they failed to give a solution to this problem. 

A straightforward idea is to collect a large number of legal cases with non-PLLS descriptions from the Web to train a prediction model. Nevertheless, this is essentially impossible. Although we could obtain many non-PLLS texts from legal forums and news Websites, the following problems hinder constructing a high-quality collection of non-PLLS fact descriptions for legal cases: (1) users engage in legal forums usually through the form of question-answering. However, few real criminal fact descriptions could be extracted therein since the stories are often secrets of the suspects (and they would not risk revealing them). Secondly, labeling the extracted fact descriptions with appropriate charges relies heavily on specialized legal knowledge and experience. (2) the accessible news reports mainly cover a few highly common charges (e.g., \emph{theft}) since they are intended for the broad public. Most charges (e.g., \emph{obstruction of witnessing}) have only a few sporadic news reports. Hence, we cannot collect enough reports for every charge from news Websites. Fortunately, we can extract the corresponding charges from these available news reports, as they paraphrase court verdicts in non-PLLS. Therefore, the best possible result is that we could collect a small number of labeled non-PLLS instances for most charges.

A possible solution to this problem is the few-shot domain adaptation (FSDA) scheme. FSDA tackles the data scarcity issue in the domain of interest (target domain) with the help of a substantial amount of labeled data in a related but differently distributed domain (source domain). FSDA methods typically try to find a shared latent space for different domains or use regularizations to improve the fit on the target domain \cite{ccsa}. We can formulate our task as an FSDA problem, where we consider fact descriptions with PLLS and non-PLLS to be the source domain and target domain, respectively. Even if some previous works such as \cite{fada,dsne,dage} could be applied to our task, they suffer from the following drawbacks:

(1) \textbf{Negative Impact of Style Information.} We have empirically observed that the main domain discrepancy for our problem lies in the different text styles, namely PLLS and non-PLLS. Existing FSDA algorithms directly align the features of data from the two domains and ignore the style information that may exist in the features, which may have a negative impact on charge prediction. Hence, it is more promising to disentangle the content and style representations for better alignment. Although disentanglement learning has been studied in unsupervised domain adaptation or text style transfer \cite{disen1,disen2,tstadv}, they assume that the target domain has a large amount of data available. Moreover, these methods mainly depend on the adversarial loss or reconstruction loss. When there is not much data in the target domain, (1) the domain/style classifier in the adversarial loss will face a severely unbalanced binary classification problem, resulting in a learning bias to the source domain; (2) the encoder-decoder network for the reconstruction loss will over-fit the target data. As a result, these disentanglement strategies are inappropriate for solving our task.

(2) \textbf{No Element-Level Alignment.} Existing FSDA methods only perform instance-level alignment, i.e., aligning the latent features of the instances belonging to the same category in different domains. For the charge prediction task, there is a legal trial principle known as \textit{Elemental Trial} for countries with a civil law system \cite{cohen1982criminal,qacharge}, which states that the legal judgments should be solely based on the crucial elements involved in the fact descriptions. Furthermore, distinguishing confusing charges also heavily depends on the crucial elements. Extracting and aligning the features in element-level are therefore critical for our task. There are mainly two types of crucial legal elements used in the literature: manually defined elements \cite{fla} and constitutive elements (CEs) of charges \cite{cecp}. The former is manually derived from law articles or judicial interpretations, such as whether the criminal has the act of violence. However, manually defining these elements requires significant expertise and effort, and is hard to be comprehensive. The latter, on the other hand, can precisely address these drawbacks. For instance, in the four-CEs framework of China, each charge is defined by its four constitutive elements, namely, \textbf{object}, \textbf{objective}, \textbf{subject}, and \textbf{subjective}. These elements collectively form a comprehensive description of a specific charge and provide discriminative information to distinguish confusing charges. It is imperative that fact descriptions match the CEs of corresponding charges.

In this paper, we propose a novel few-shot domain adaptation method for predicting charges with non-PLLS fact descriptions, named Disentangled Legal Content for Charge Prediction (DLCCP). Similar to existing FSDA methods, DLCCP employs the pairing strategy to align the target data with all source data to address the data scarcity issue in the target domain. The key difference is, we disentangle fact descriptions (also CEs) into content representation (CR)\footnote{Both instance-level and element-level CRs are learned. The different types of CEs of a charge are processed sequentially using a GRU to obtain the instance-level CR.} and style representation (SR) to reduce the domain discrepancy due to expressive styles. For the CRs from two domains, which can be regarded as domain-invariant representations, we propose to achieve intra-class alignments in both element-level and instance-level, and also inter-class separation in instance-level. We do not perform element-level inter-class separation since there may be some similar elements for different charges. DLCCP also matches the element-level and instance-level CRs from two domains with the CRs of corresponding CEs to encourage the disentangled CRs to capture discriminative information for different charges. For SRs, we impose a clustering constraint to force SRs in the same domain to be close, since they represent the same text style. Two additional regularizations are performed at corpus-level by taking the average of SRs within a domain or over all charges' CEs: (1) separation between the two domains; (2) alignment between the source data and CEs since they are both written by legal experts. These carefully designed alignment, separation and clustering losses empower the learning of domain-invariant and semantically meaningful CRs, as well as domain-dependent and semantically absent SRs. The incorporation of the prior knowledge of CEs provides a valuable guide for content/style disentanglement, which would lead to much less required training data compared to previous disentanglement learning \cite{disen1,disen2,tstadv}.

In summary, the contributions of this work are: (1) to the best of our knowledge, we provide the first solution to tackle the domain diversity between PLLS and non-PLLS fact descriptions for charge prediction; (2) a novel few-shot domain adaptation method is proposed to perform element-level, instance-level and corpus-level alignment or separation simultaneously to learn better legal-specific and domain-invariant CRs; (3) we contribute a non-PLLS dataset called News Corpus for Charge Prediction (NCCP). And we conduct plentiful experiments on NCCP to verify the effectiveness of our model and show its significant improvements over competitive baselines.

\section{Related Work}

\subsection{Charge Prediction}

\noindent The automatic prediction of charges based on fact descriptions has been investigated as a text classification problem for decades. Early studies relied mainly on manually designed legal-specific features or mathematical methods \cite{mathematical-1,mathematical-2} because there was a paucity of data. However, these works exhibit substantial reliance on manual efforts, making it challenging to apply to diverse scenarios. In recent years, the availability of numerous large-scale datasets \cite{surveyLJP} has made neural network-based methods the dominant approach for charge prediction. These works may generally be separated into the following four groups. 

Firstly, some studies focus on coping with the textual characteristics of fact descriptions, primarily longer text length than ordinary texts. \cite{sattcaps} devised a self-attentive dynamic routing mechanism for capsule network to capture long-range dependency in fact descriptions. \cite{longtransformer} released the Longformer-based pre-trained model for long legal documents. 
Secondly, some works explicitly incorporate legal knowledge to support in the extraction of legal-specific features to enhance the performance. \cite{fewshot} introduced a set of discriminative attributes of charges serving as the internal mapping between charges and fact descriptions, and proposed an attribute-attentive network for charge prediction. \cite{ladan} proposed a novel graph neural network for the automatic learning of subtle distinctions among law articles, which were taken to attentively extract discriminative features from fact descriptions. \cite{cecp} and \cite{inprovingljp} explored the matching between fact descriptions and CEs of charges, and \cite{hierarchicalmatching} and \cite{leapbank} explored the matching between fact descriptions and charges.  
Thirdly, some studies explore the correlation between the charge prediction task and other legal judgment prediction tasks, such as law article recommendation and prison term prediction. These works are commonly structured within the framework of multi-task learning (MTL). Unlike conventional MTL models, which primarily focus on parameter sharing among tasks, these works typically exploit the dependencies among legal subtasks. \cite{topjudge} utilized a directed acyclic graph to capture the topological dependencies among legal subtasks. \cite{bifeedback} designed a backward verification mechanism to leverage the interdependencies of results among legal subtasks. \cite{mlljp} introduced distinct role embeddings for charge-related law articles and prison term-related law articles to model the dependencies between different types of articles and terms. 
All three of the aforementioned groups are strongly dedicated to the goal of minimizing the prediction error rate.
Finally, other studies concentrate on the interpretability of predictions. \cite{qacharge} proposed to iteratively ask questions to detect key elements in the fact descriptions and the response could be used to analyze the prediction results. \cite{double_layer} designed a double-layer criminal system to extract objective elements and subjective elements from fact descriptions for interpretable charge prediction. 

All of these existing efforts, however, rely on PLLS texts as input. Such a learned model can only be advantageous to legal experts. Despite the fact that our work falls within the second category, we concentrate on non-PLLS texts to make the charge prediction model accessible to laypersons.

\subsection{Domain Adaptation}
Domain adaptation (DA) aims to handle the domain discrepancy between the source domain and the target domain. The current DA methods may be categorized into three groups depending on the characteristics of the target domain data: unsupervised DA (UDA), semi-supervised DA (SSDA), and supervised DA (SDA). UDA and SSDA presume that the target domain has access to a large amount of unlabeled data \cite{uda}. On the contrary, the SDA works on the case where extremely few labeled target samples are accessible, and is also called FSDA. Due to the previously mentioned difficulties in collecting numerous labeled or unlabeled fact descriptions with non-PLLS, the FSDA methods are more related to our work. 
\cite{fada} exploited the adversarial learning to maximize the domain confusion and semantically align features. In \cite{dsne}, the stochastic neighborhood embedding technique was employed to maximize the inter-class distance and minimize the intra-class distance for both domains. \cite{dage} proposed to learn domain-invariant features by multi-view graph embedding. \cite{karanam2022orient} employed the submodular mutual information function to carefully select a subset of source data, thus enhancing the training efficiency.
However, the features learned and aligned by these methods may contain style information, and these works cannot perform element-level alignment which is crucial for the charge prediction task. These factors make these works may not work well on our task. There are also other FSDA methods designed for specific fields such as pedestrian detection \cite{PedestrianDetec} and regression task \cite{regression}, which cannot be applied to our problem. 

On the other hand, disentanglement learning has been well investigated within the realm of UDA \cite{disen1,disen2,disen3}, which primarily involves the domain classifier or encoder-decoder network to provide supervision signal for learning domain-invariant and domain-dependent features. However, as previously mentioned, applying these disentanglement techniques to FSDA poses challenges as the domain classifier will learn bias toward the source domain and the encoder-decoder network will suffer from the over-fitting issue when confronted with data scarcity in the target domain.

\subsection{Text Style Transfer}

Text style transfer is a methodology for automatically generating text with a certain style (e.g., formality and humor) while maintaining the content of the original text. At present, the majority of research is based on non-parallel data since obtaining parallel data is difficult and challenging in many applications, such as topic transfer \cite{tstsurvey1}. These approaches can be roughly categorized into non-disentanglement methods and disentanglement-based methods. And techniques employed include prototype editing \cite{prototypeedit} and pseudo-parallel corpus construction \cite{pseudoparallel}. The former does not disentangle the representations of content and style, whereas the latter does and relies primarily on adversarial loss \cite{tstadv} or cycle loss \cite{tstcircle}. We embrace the fundamental principle of the latter and disentangle texts into CRs and SRs to mitigate the domain discrepancy (i.e., PLLS and non-PLLS) within our task. However, similar to the disentanglement learning in UDA, these works also rely on the style classifier or the encoder-decoder network and suffer from learning bias or over-fitting issues.

\section{Method}

\begin{figure*}[t]
    \centering
    \includegraphics[width=0.85\textwidth]{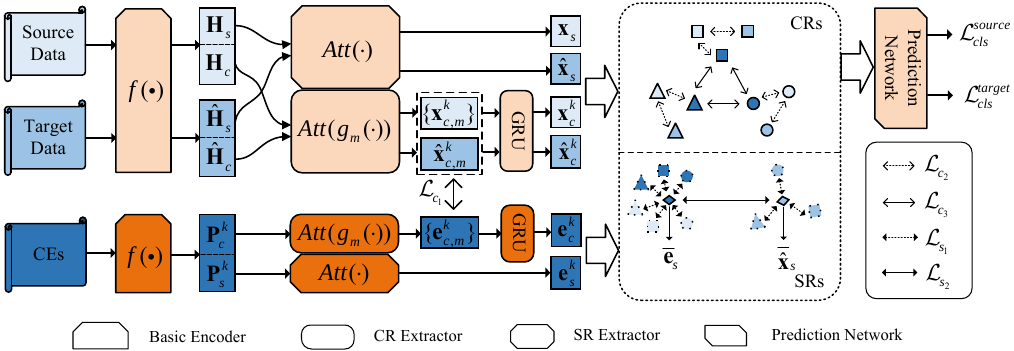}
    \caption{The Framework of DLCCP. Varying shades of blue indicate distinct types of data, while different shades of orange represent the networks used for fact descriptions or CEs. On the right-hand side, the squares, triangles, and circles respectively correspond to CRs (instance-level) or SRs belonging to three distinct charges.}
    \label{fig:framework}
\end{figure*}

\subsection{Problem Formulation}

\noindent A legal case contains a fact description and the corresponding charge. We denote the source data with $\lvert \mathcal{S} \rvert$ samples as $\mathcal{S} = \{X_i, y_i \}_{i=1}^{\lvert \mathcal{S} \rvert}$, where $X_i = \{w_j \}_{j=1}^{n_x}$ refers to the $i$-th sample with $n_x$ words and $y_i$ is the corresponding charge. Similarly, the target data with $\lvert \mathcal{T} \rvert$ samples is denoted as $\mathcal{T} = \{\hat{X}_i, \hat{y}_i\}_{i=1}^{\lvert \mathcal{T} \rvert}$. Since we focus on the FSDA scenario, we have $\lvert \mathcal{T} \rvert \ll \lvert \mathcal{S} \rvert$. It should be noted that the two domains share a common label space with a total of $K$ charges, i.e., $y_i, \hat{y}_i \in \{1, \dots, K\}$. 
Additionally, we have $M$ types CE knowledge for each charge, each of which contains $n_m$ words and is represented as $E_{m}^{k} = \{w_j \}_{j=1}^{n_m}$. We use $E^k$ to denote all CEs $\{ E_m^k \}_{m=1}^M$ for a charge $k$. In the following, we exclude the sample's indicator $i$ for the purpose of conciseness and we use bold face lower/upper case letters to denote vectors/matrices respectively. We summarize the main notations used in this paper in Table \ref{tab:notations} to provide a clearer depiction of them. Based on $\mathcal{S}$ and $\mathcal{T}$, as well as $\{ E^{k} \}_{k=1}^K$, our goal is to develop a model that can precisely predict the corresponding charge $\hat{y}$ for a given $\hat{X}$. 

\subsection{DLCCP}

In DLCCP, we disentangle the representation of the fact description (also the CE) into CR and SR, and predict the corresponding charge based on the instance-level domain-invariant CR. As shown in Fig. \ref{fig:framework}, the proposed method mainly consists of \textbf{basic encoder}, \textbf{CR extractor}, \textbf{SR extractor}, and \textbf{prediction network}, which are represented as rectangles with distinct vertices. Initially, the basic coder is responsible for capturing the dependencies among words and outputs a mixed representation $\mathbf{h}$ for each word. Then, we separate the $\mathbf{h}$ into $\mathbf{h}_c$ and $\mathbf{h}_s$, which are passed to the CR extractor and SR extractor, respectively. The CR extractor will learn $M$ element-level CRs, which correspond to the $M$ CE types of a charge, and an instance-level CR for $X$, $\hat{X}$, or $E^{k}$. Note that the element-level CRs are not used to predict the charge, but serve to learn a better instance-level CR. On the other hand, the SR extractor learns a unique SR for $X$, $\hat{X}$, or $E^{k}$. Finally, the instance-level CR of a fact description is fed to the prediction network to get the charge. Inspired by \cite{fada,dsne}, we employ the pairing strategy of source data and target data to address the data scarcity issue of the target domain. But in DLCCP, we actually pair two triples (i.e., $(X^{k_1}, \hat{X}^{k_1}, E^{k_1})$ and $(X^{k_2}, \hat{X}^{k_2}, E^{k_2})$, where $k_1 \neq k_2$). In the following, we will elaborate on the four components of DLCCP.

\begin{table}
    \tabcolsep=0.14cm
	\caption{Main Notations.}
	\label{tab:notations}
        \centering
	\renewcommand{\arraystretch}{1.3} 
	\begin{tabular}{l p{6.6cm}}
		\hline
		Notations & Descriptions\\
		\hline
  
		$m$ & Indicator of CE type\\
            $M$ & Total number of CE types\\
		$k$ & Indicator of charge category\\
            $K$ & Total number of charge categories\\
            $X/\hat{X}$ & Fact descriptions from the source/target domain \\
		$E^k$ & Set of $m$ CEs of charge $k$ \\
            $\mathbf{x}_c/\mathbf{\hat{x}}_{c}$ & Instance-level CRs of fact descriptions from the source/target domain \\
            $\mathbf{x}_{c,m}/\mathbf{\hat{x}}_{c,m}$ & Element-level CRs of fact descriptions from the source/target domain \\
            $\mathbf{x}_s/\mathbf{\hat{x}}_s$ & SRs of fact descriptions from the source/target domain \\
            $\mathbf{e}_c/\mathbf{e}_{c,m}$ & Instance-level/element-level CRs of CEs \\
            $\mathbf{e}_{s,m}^k$ & SR of type $m$ CE  of charge $k$ \\
  
		\hline
	\end{tabular}
\end{table}

\subsubsection{Basic Encoder}

The original discrete words are converted into $d$-dimensional continuous embeddings via the basic encoder. Here, our model is independent of the basic encoder and it can be any neural network (e.g., the RNN-based and Transformer-based network) that can capture the dependencies among words and output a sequence of hidden representations. Formally, take the source data $X$ as an example and the basic encoder can be denoted as:
\begin{equation}
\label{eq:basic_encoder}
    \mathbf{H} = f(X),
\end{equation}
where $\mathbf{H} = [\mathbf{h}_1, \mathbf{h}_2, \dots, \mathbf{h}_{n_x}] \in \mathbb{R}^{d \times n_x}$. Inspired by \cite{tstadv}, we split each hidden state $\mathbf{h}_i$ ($i=[1,2,\dots,n_x]$) into $\mathbf{h}_{ic} \in \mathbb{R}^{d_c}$ and $\mathbf{h}_{is} \in \mathbb{R}^{d_s}$ (i.e., $\mathbf{h}_i = [\mathbf{h}_{ic}; \mathbf{h}_{is}]$), which contain the content information and style information for each word, respectively. Hence we obtain $\mathbf{H}_c \in \mathbb{R}^{d_c \times n_x}$ and $\mathbf{H}_s \in \mathbb{R}^{d_s \times n_x}$ to be passed to the CR extractor and the SR extractor, respectively. We employ the same process to obtain $\hat{\mathbf{H}}_c$ and $\hat{\mathbf{H}}_s$ for the target data $\hat{X}$, and $\mathbf{P}^k_c = \{ \mathbf{P}_{c,m}^{k} \}_{m=1}^{M}$ and $\mathbf{P}_s^k = \{ \mathbf{P}_{s,m}^{k} \}_{m=1}^{M}$ for the CEs $E^k$. For the basic encoder and the following modules: the network parameters are shared for the source data and target data, and we use independent network parameters for CEs since their text lengths are different from fact descriptions.

\subsubsection{CR Extractor} 
The CR extractor takes $\mathbf{H}_c$, $\hat{\mathbf{H}}_c$ and $\mathbf{P}_{c}^{k}$ as input and performs alignments in both element-level and instance-level, as well as separation in instance-level. We then elucidate these mechanisms, along with their respective functionalities.

First, we learn element-level CRs. Also, take the source data $X$ as an example. We use $M$ non-linear network $g_m$ to transform the $\mathbf{H}_c$ to $M$ latent spaces, each of which is associated with a corresponding CE type (in our experimental section, we employ a single fully connected layer followed by a Tanh activation function for $g_m$). To capture the most important words concerning a certain CE type, we employ the attention mechanism ($Att$) to select the relevant information. Formally, the above process can be formulated as:
\begin{equation}
\label{eq:att}
\begin{split}
    \mathbf{h}_{c,i,m} &= g_m(\mathbf{H}_{c, i}), \\
    {\alpha}_{c,i,m} &= \frac {exp((\mathbf{u}_{c,m})^{\intercal} \mathbf{h}_{c,i,m})} {\sum\nolimits_{i=1}^{n_x} exp((\mathbf{u}_{c,m})^{\intercal} \mathbf{h}_{c,i,m})}, \\
    \mathbf{x}_{c,m} &= \sum\nolimits_{i=1}^{n_x} {\alpha}_{c,i,m} \mathbf{h}_{c,i,m},
\end{split}
\end{equation}
where $\mathbf{H}_{c,i}$ is the $i$-column of $\mathbf{H}_c$ and $\mathbf{u}_{c,m}$ represents a learnable query vector used to assess the informativeness of the CR of a word with respect to the current $m$-th CE type. Thus we obtain the element-level CRs denoted as $\{ \mathbf{x}_{c,m} \}_{m=1}^M$ for an input fact description.

Second, we aim to capture the knowledge of logical order among these element-level CRs, which follows the human’s charge identification process in practice \cite{logicalorder,cecp}. Furthermore, we will learn an instance-level CR $\mathbf{x}_c$ derived from element-level CRs for a fact description. Concretely, we construct the $\{\mathbf{x}_{c,m} \}_{m=1}^{M}$ into $Seq = [\mathbf{x}_{c,1}, \mathbf{x}_{c,2}, \dots, \mathbf{x}_{c,M}]$, and the position of $\mathbf{x}_{c,m}$ in $Seq$ matches the logical order among different CE types. Then we feed the $Seq$ into a GRU \cite{gru} network and the final hidden state for the input sequence is regarded as the instance-level CR of a fact description. Formally, the entire two steps can be formulated as:
\begin{equation}
\label{eq:content}
    \mathbf{x}_c = GRU( [Att(g_m(\mathbf{H}_c))]_{m=1}^M )_{final \ stste}.
\end{equation}

In the same process, we can get the element-level CRs $\{ \hat{\mathbf{x}}_{c,m} \}_{m=1}^M$ and instance-level CR $\hat{\mathbf{x}}_c$ for the target data $\hat{X}$. For CEs, the only difference is that the input of $g_m$ is $\mathbf{P}_{c,m}^k$. And we also get the element-level CRs $\{ \mathbf{e}_{c,m}^k \}_{m=1}^M$ and instance-level CR $\mathbf{e}_c^k$ for the CEs $E^k$.

In the content space, the semantic alignments in element-level and instance level are defined as:
\begin{equation}
\begin{split}
    \mathcal{L}_{c_1} &= \sum\nolimits_{m=1}^{M} d(\mathbf{x}_{c,m}^k, \hat{\mathbf{x}}_{c,m}^k) + d(\mathbf{x}_{c,m}^k, \mathbf{e}_{c,m}^k), \\
    \mathcal{L}_{c_2} &= d(\mathbf{x}_c^{k}, \hat{\mathbf{x}}_c^{k}) + d(\mathbf{x}_c^{k}, \mathbf{e}_c^{k}), 
\end{split}
\end{equation}
where $d$ is a similarity metric function and we use the Euclidean distance for it in our experiments. Here we add a superscript $k$ to indicate the label of CRs of fact descriptions. We make these CRs from different domains with the same class to be aligned semantically, which enables the learning of domain-invariant CRs. In this way, it is hard to judge whether the CR of a sample comes from the source domain or the target domain. At the same time, they are aligned with the CRs of CEs of the corresponding charge to enable the extracting of legal-specific CRs and encourage the capture of discriminative information to distinguish confusing charges. Here the alignment between $\hat{\mathbf{x}}_{c,m}^k$ ($\hat{\mathbf{x}}_c^k$) and $\mathbf{e}_{c,m}^k$ ($\mathbf{e}_c^k$) is transmitted by $\mathbf{x}_{c,m}^k$ ($\mathbf{x}_c^k$). The instance-level separation is defined as:
\begin{equation}
    \mathcal{L}_{c_3} = exp(- d (\mathbf{e}_c^{k_1}, \mathbf{e}_c^{k_2})),
\end{equation}
where $k_1$ and $k_2$ indicate two different charges. In $\mathcal{L}_{c_3}$, we maximize the similarity of instance-level CRs of CEs of different charges. The constraints $\mathcal{L}_{c_2}$ and $\mathcal{L}_{c_3}$ enable the instance-level CRs with the same charge from different domains to converge into a single cluster while maximizing the separation between clusters associated with distinct charges. Since there may be some similar elements in different charges, element-level separation for different charges is not in accordance with the laws and regulations, and we ignore this separation.

Finally, the total loss for CRs is:
\begin{equation}
    \mathcal{L}_c = \lambda_{c_1} \mathcal{L}_{c_1} + \lambda_{c_2} \mathcal{L}_{c_2} + \lambda_{c_3} \mathcal{L}_{c_3}.
\end{equation}

\subsubsection{SR Extractor} 
The SR extractor takes $\mathbf{H}_s$, $\hat{\mathbf{H}}_s$ and $\mathbf{P}_{s}^k$ as input and performs with-domain clustering, as well as alignment and separation in corpus-level. For natural language, the SR is a corpus-level variable \cite{corpus-feature}. Therefore, the SRs corresponding to different samples in the same domain should be similar. From this point of view, we cluster the SRs of different domains into separate clusters.

Some words do not flow well into text style \cite{contentenhancing}. Hence, we employ the attention mechanism to select style-related information attentively. Formally, this process can be represented as (take $\mathbf{H}_s$ of the source data as an example):

\begin{equation}
    \mathbf{x}_s = Att( \mathbf{H}_s ),
\end{equation}
where the $Att$ has the same structure as the $Att$ in CR extractor, but with different parameters. In the same way, we can obtain $\hat{\mathbf{x}}_s$ for the target data $\hat{X}$ and $\{ \mathbf{e}_{s,m}^k \}_{m=1}^{M}$ for the CEs. And we define $\mathbf{e}_s^k = \frac{1}{M} \sum\nolimits_{m=1}^M \mathbf{e}_{s,m}^k$ to be the SR for $E^k$.

Let $\bar{\mathbf{x}}_s = \frac{1}{bs} \sum\nolimits \mathbf{x}_{s}$ and $\bar{\hat{\mathbf{x}}}_s = \frac{1}{bs} \sum\nolimits \hat{\mathbf{x}}_s$ be the mini-batch-based centers of SRs for the source and target data respectively, where $bs$ is the size of mini-batch. We define the $\mathcal{L}_{s_1}$ to make the SRs of the same domain (as well as CEs) clustered. On the other hand, since the source data and the CEs are both written by legal experts (i.e., they both exhibit the PLLS), we align the SRs of them. Here we align the SRs of the CEs with $\bar{\mathbf{x}}_s$, which can simultaneously enable the clustering within the CEs and the corpus-level alignment between the source data and the CEs. The complete $\mathcal{L}_{s_1}$ can be formulated as:
\begin{equation}
    \mathcal{L}_{s_1} =  d(\hat{\mathbf{x}}_s, \bar{\hat{\mathbf{x}}}_s) + d(\mathbf{x}_s, \bar{\mathbf{x}}_s) + d(\mathbf{e}_s^k, \bar{\mathbf{x}}_s).
\end{equation}

For the corpus-level SRs $\bar{\mathbf{x}}_s$ and $\bar{\hat{\mathbf{x}}}_s$ of the source and target data, we introduce the following constraint to make these two centers move away from each other:
\begin{equation}
    \mathcal{L}_{s_2} = exp(-d (\bar{\mathbf{x}}_s, \bar{\hat{\mathbf{x}}}_s)).
\end{equation}

Finally, the total loss for SRs is:
\begin{equation}
    \mathcal{L}_s = \lambda_{s_1} \mathcal{L}_{s_1} + \lambda_{s_2} \mathcal{L}_{s_2}.
\end{equation}

\subsubsection{Prediction Network}
Based on the disentangled and GRU-derived instance-level CRs, we employ a linear classifier with parameters $\mathbf{W}$ and $\mathbf{b}$, followed by a softmax activation function, to calculate the probability distribution of charges:

\begin{equation}
    \widetilde{\mathbf{y}} = \operatorname{softmax} (\mathbf{W} \mathbf{c} + \mathbf{b}), \quad \quad \text{for} \ \mathbf{c} = \mathbf{x}_c \ \text{or} \ \hat{\mathbf{x}}_c.
\end{equation}

Since we focus on a supervised scenario, the classification losses for the source and target data can also be exploited. We additionally employ stand cross-entropy losses $\mathcal{L}_{cls}^{source}$ and $\mathcal{L}_{cls}^{target}$ for the source and target data respectively. The total loss for DLCCP is:
\begin{equation}
\label{eq:total_loss}
    \mathcal{L} = \mathcal{L}_c + \mathcal{L}_s + \beta_1 \mathcal{L}_{cls}^{source} + \beta_2 \mathcal{L}_{cls}^{target}.
\end{equation}

\subsection{Analysis about the Disentanglement in DLCCP.}

In the literature \cite{tstadv,contentenhancing,tstsurvey2}, text content is typically considered domain-invariant, while text style tends to be domain-dependent. Furthermore, the former should be capable of conveying the essence and semantic meaning of the text, whereas the latter does not achieve this in any way. 
To disentangle CR and SR in DLCCP, we impose: (1) inter-domain alignment to align CRs with the same charge and intra-domain separation to separate CRs with different charges. Specifically, the features (CRs) from two domains will be aligned based on their label information when minimizing $\mathcal{L}_{c_1}$ and $\mathcal{L}_{c_2}$, and the instance-level features (CRs) from a single domain will be separated based on their label information when minimizing $\mathcal{L}_{c_2}$ and $\mathcal{L}_{c_3}$. In this way, the learned CRs will not change between domains and can capture the fundamental aspects of texts to distinguish charges. (2) inter-domain separation to separate SRs of two domains and intra-domain alignment to align SRs with the same domain. In detail, the features (SRs) from two domains will be separated when minimizing $\mathcal{L}_{s_2}$, and the features (SRs) from a single domain will be clustered when minimizing $\mathcal{L}_{s_1}$. Hence, the learned SRs exhibit variations across domains even for texts describing the same crime, but are similar within the same domain, leading to a loss in their ability to depict the factual details of the crime.
Consequently, the CRs and SRs are disentangled naturally and intuitively. This mechanism allows us to avoid the earn bias and over-fitting issues existing in the traditional disentanglement strategies within the context of UDA.

\section{Experiments}
\noindent In this section, we conduct extensive experiments to demonstrate the performance of DLCCP. Our methods and baselines are implemented by PyTorch and run on Ubuntu Linux 20.04. (Hardware used: 32-core AMD EPYC CPU 7543 at 2.80 GHz, NVIDIA GeForce RTX 3090). An implementation of the proposed method and the dataset used in our experiments are available online\footnote{\url{https://github.com/jiezhao6/DLCCP}}.

\subsection{Datasets}

As previously discussed, existing studies on charge prediction have focused on fact descriptions with PLLS. Consequently, the existing datasets are also composed of PLLS texts obtained from publicly available court judgments. To evaluate the performance of DLCCP, we employ two widely-used datasets, namely Criminal \cite{fewshot} and CAIL \cite{cail}, as the source data. Additionally, we collect a non-PLLS target dataset named News Corpus for Charge Prediction (\textbf{NCCP}). For the knowledge of CEs, we utilize the publicly available data released by \cite{cecp}. Next, we provide detailed introductions to these data and some preprocessing steps we performed on them.

\paragraph{Source Data} Criminal and CAIL are both collected from the China Judgments Online\footnote{\url{https://wenshu.court.gov.cn/}}. The Criminal consists of roughly 500,000 legal cases, each containing a fact description and the corresponding charge. Three sub-datasets with varying sizes, namely Criminal-S, Criminal-M, and Criminal-L, are constructed by partitioning the main Criminal dataset. 
CAIL is a multi-label dataset, consisting of an extensive collection of over 1.68 million legal cases. It has also been partitioned into sub-datasets with different sizes, known as CAIL-Small and CAIL-Big. We follow the preprocessing steps proposed in \cite{ladan,cecp} to prepare the two sub-datasets. Specifically, we extract the fact descriptions and charges and filter out the cases with multi-charges. We conduct further filtering by excluding cases with fact descriptions containing fewer than 30 words, as such short texts are found to lack sufficient information relevant to charge prediction.
Important statistics of Criminal and preprocessed CAIL are presented in Table \ref{exp:data-statistic-source} (the numbers outside the parentheses).

\paragraph{Target Data} There is no publicly available non-PLLS dataset for charge prediction in previous works. Considering the Chinese judgment dataset is used as the source data in our experiments, we collect non-PLLS data from the news website of CCTV\footnote{\url{https://www.cctv.com/}} automatically. In order to fit our FSDA task, we filter out these charges with fewer than 10 cases and samples with multiple charges. Finally, we obtain a target dataset named NCCP. Important statistics of NCCP are summarized in Table \ref{exp:data-statistic-target}.

\paragraph{CEs Data} Zhao \textit{et al.} \cite{cecp} collected CEs descriptions of charges from zuiming.net\footnote{\url{https://www.zuiming.net/}}. Each charge is associated with four CE types, namely subject element, subjective element, object element, and objective element. And each element is composed of several paragraphs with PLLS text.

\begin{table}[!t]
        \caption{Statistics of the source datasets. The numbers in parentheses indicate the case and charge counts after removing a few cases to maintain the consistency of label space.}
	\label{exp:data-statistic-source}
	\centering
	\begin{tabular}{lrl} \toprule
		Statistics & Training Cases & Charges \\
		\midrule
		  Criminal-S & 61,589 (61,150) & 149 (129) \\
            Criminal-M & 153,521 (152,583) & 149 (129) \\
            Criminal-L & 306,900 (305,067) & 149 (129) \\
            CAIL-Small & 101,275 (84,732) & 119 (93) \\
            CAIL-Big   & 1,585,765 (1,521,942) & 128 (98) \\
		\bottomrule
	\end{tabular}
\end{table}

\begin{table}[!t]
        \caption{Statistics of the NCCP dataset.}
	\label{exp:data-statistic-target}
	\centering
	\begin{tabular}{ccccc} \toprule
		Statistics & Training Cases & Valid Cases & Test Cases & Charges \\
		\midrule
		  NCCP & 774 & 3,539  & 3,603  & 129\\ 
		\bottomrule
	\end{tabular}
\end{table}

There are some uncommon charges for which fact descriptions are difficult to acquire, even when focusing on the FSDA scenario. To maintain a consistent label space across both domains, we further apply some filtration steps to these datasets. Specifically : (1) when considering Criminal as the source domain, we remove 20 charges that are not present in NCCP from Criminal, along with their corresponding fact descriptions; (2) when using CAIL as the source domain, we retain only the cases whose corresponding charges are present in both domains. The numbers in parentheses in Table \ref{exp:data-statistic-source} represent the statistics of the source data after this filtering process.

\subsection{Baselines}

There is no prior work focusing on charge prediction across different domains and we organize the following three types of baselines.

\begin{itemize}
\item{We first employ these methods that are designed for the single domain charge prediction task (i.e., using fact descriptions with PLLS as inputs). \textbf{FewShot} \cite{fewshot} extracts manually defined legal elements from fact descriptions and is an element-attentive charge prediction model. \textbf{CECP} \cite{cecp} is a reinforcement model where an agent selects crucial sentences as instances of CEs based on the matching between CEs and fact descriptions. \textbf{BERTL} \cite{bert-pretrain} is pre-trained on large-scale legal texts. And \textbf{LeapBank} \cite{leapbank} explores the label information by fusing the representations of fact descriptions and texture content of labels.}

\item{We train the aforementioned methods on the hybrid data and denote these baselines as ``\textbf{\textit{method}+T}''. This signifies that we mix the training samples from both the source and target domains and utilize them equally during the training of the first group of baselines.}

\item{The third group involves some FSDA methods that can be applied to our task. \textbf{FADA} \cite{fada} employs adversarial learning to extract domain-invariant features. \textbf{$d$-SNE} \cite{dsne} and \textbf{DAGE} \cite{dage} utilize the stochastic neighborhood embedding and graph embedding to align features from different domains respectively.}

\end{itemize}

Note that the BERTL and LeapBank are transformer-based methods. To have a fair comparison with all baselines, we employ the GRU \cite{gru} or the pre-trained BERT model from HuggingFace\footnote{\url{https://huggingface.co}} as the basic encoder, denoted as \textbf{DLCCP+GRU} and \textbf{DLCCP+BERT}, respectively. For FADA, $d$-SNE, and DAGE, we use the GRU as the encoder to compare them with DLCCP+GRU. We adopt the settings from the original papers for other methods.

\subsection{Hyperparameters Settings}

The maximum document length of fact descriptions is set to 512  words for all models except CECP. CECP employs a hierarchical structure encoder and we set the maximum sentence length as 32 words and the maximum document length as 64 sentences. Within this setting, the average word count contained in the fact descriptions is nearly consistent across all models. For CEs, we follow the setting of \cite{cecp} and set the maximum length of subject element, subjective element, object element, and objective element as 100, 100, 200, and 400 words, respectively. And we set the dimensionality of CR and SR as 128 and 16 for DLCCP-GRU, and 704 and 64 for DLCCP-BERT. For other RNN-based and transformer-based models, we respectively set the latent states to 128 and 768.

For the other hyperparameters in DLCCP, we select them according to the performance on the validation set in NCCP. Specifically, the parameters $\lambda_{c_1}$, $\lambda_{c_2}$, $\lambda_{c_3}$, $\lambda_{s_1}$, $\lambda_{s_2}$, $\beta_1$ and $\beta_2$ are set to 1, 1, 10, 1, 1, 100 and 1, respectively. For the other hyperparameters in baselines, we follow the settings described in the original papers.

\subsection{Training Details}

To pre-train word embeddings for those RNN-based models, we employ THULAC\footnote{\url{http://thulac.thunlp.org/}} for word segmentation due to the Chinese being used in the raw data. Subsequently, we train a Skip-Gram model on the segmented data with a frequency threshold of 25 to obtain word embeddings. And the word embedding size is set to 200. For all models, we employ the Adam optimizer to optimize them. We use learning rates of $1 \times 10^{-3}$ and $2 \times 10^{-5}$ for RNN-based models and transformer-based models, respectively. The batch sizes for these two types of models are set as 64 and 16. 

We empirically find that the cross entropy losses of $\mathcal{L}_{cls}^{source}$ and $\mathcal{L}_{cls}^{target}$ decrease quite slowly if we directly employ the entire loss $\mathcal{L}$ (defined in Eq.~(\ref{eq:total_loss})) to train the DLCCP. This could be ascribed to the challenge of simultaneously satisfying both disentanglement and classification constraints within the hidden space associated with randomly initialized parameters. Therefore, we perform a pre-training process where we employ the loss of $\beta_1\mathcal{L}_{cls}^{source} + \beta_2 \mathcal{L}_{cls}^{target} + \lambda_{c_3}\mathcal{L}_{c_3}$ to make the model focus on classifying fact descriptions and separating CEs of different charges. Subsequently, we train the DLCCP with the total loss $\mathcal{L}$. The pre-training epochs (the subsequent training epochs) are set to 20 and 2 (20 and 3) for DLCCP+GRU and DLCCP+BERT, respectively.

\subsection{Metrics}

\begin{table*}[!t]
        \caption{Results under the one-shot setting using Criminal as the source data. Underlined values denote the optimal results of baselines.}
	\label{exp:oneshot}
	\centering
	\begin{tabular}{lcccc|cccc|cccc} \toprule
		Datasets  
		& \multicolumn{4}{c}{\textbf{Criminal-S} $\rightarrow$ \textbf{NCCP}}
		& \multicolumn{4}{c}{\textbf{Criminal-M} $\rightarrow$ \textbf{NCCP}} 
		& \multicolumn{4}{c}{\textbf{Criminal-L} $\rightarrow$ \textbf{NCCP}} \\
		\cmidrule(lr){2-5} \cmidrule(lr){6-9} \cmidrule(lr){10-13}
		Metrics  & Acc.  & MP    & MR    & F1    & Acc.  & MP    & MR    & F1    & Acc.  & MP    & MR    & F1 \\
		\midrule
            FewShot
            & 0.5154 & 0.4203 & 0.4000 & 0.3609 
		& 0.5543 & 0.4748 & 0.4394 & 0.4176 
		& 0.5722 & \underline{0.5014} & 0.4676 & 0.4343 \\
		CECP  
		& 0.5012 & 0.4251 & 0.3667 & 0.3549 
		& 0.5401 & 0.4702 & 0.4219 & 0.4056 
		& 0.5537 & 0.4797 & 0.4407 & 0.4191 \\
            FewShot+T
		& 0.5538 & 0.4394 & 0.4538 & 0.4044 
		& 0.5709 & 0.4819 & 0.4652 & 0.4250 
		& 0.5925 & 0.4971 & 0.4833 & 0.4464 \\
            CECP+T
		& 0.5307 & 0.4314 & 0.4146 & 0.3840 
		& 0.5514 & 0.4642 & 0.4416 & 0.4135 
		& 0.5601 & 0.4999 & 0.4911 & 0.4562 \\
            \midrule
		  FADA  
		& 0.5739 & \underline{0.4758} & \underline{0.5130} & \underline{0.4467} 
		& \underline{0.5877} & \underline{0.4886} & \underline{0.5443} & \underline{0.4700} 
		& \underline{0.5949} & 0.5003 & \underline{0.5691} & \underline{0.4816} \\
            $d$-SNE  
		& \underline{0.5745} & 0.4735 & 0.4948 & 0.4461 
		& 0.5153 & 0.4404 & 0.4710 & 0.4105 
		& 0.5509 & 0.4635 & 0.4971 & 0.4344 \\
            DAGE  
		& 0.5337 & 0.4513 & 0.4710 & 0.4149 
		& 0.5714 & 0.4758 & 0.5014 & 0.4448 
		& 0.5720 & 0.4795 & 0.5137 & 0.4531 \\

		\midrule
		\textbf{DLCCP+GRU}  
		& \textbf{0.5795} & \textbf{0.4775} & \textbf{0.5342} & \textbf{0.4679} 
		& \textbf{0.6008} & \textbf{0.5056} & \textbf{0.5604} & \textbf{0.4969} 
		& \textbf{0.6045} & \textbf{0.5052} & \textbf{0.5787} & \textbf{0.4972} \\

            \midrule
            BERTL  
		& 0.5673 & 0.4253 & 0.4180 & 0.3881 
		& 0.5390 & 0.4723 & 0.4144 & 0.4036 
		& 0.5726 & 0.5164 & 0.4540 & 0.4437 \\
		  LeapBank
		& 0.5871 & \underline{0.5331} & 0.4989 & 0.4712 
		& 0.6071 & \underline{\textbf{0.5685}} & 0.5085 & \underline{0.4993} 
		& 0.6019 & 0.5637 & 0.5006 & 0.4853 \\
            RERTZoo+T
		& 0.5545 & 0.4048 & 0.4286 & 0.3764 
		& \underline{0.6114} & 0.4802 & 0.4665 & 0.4399 
		& 0.5931 & 0.5130 & 0.4914 & 0.4602 \\
		  LeapBank+T
		& \underline{0.6241} & 0.5054 & \underline{0.5212} & \underline{0.4778} 
		& 0.6112 & 0.5449 & \underline{0.5265} & 0.4955 
		& \underline{0.6268} & \underline{\textbf{0.5901}} & \underline{0.5174} & \underline{0.5107} \\
            \midrule
            \textbf{DLCCP+BERT}  
		& \textbf{0.6282} & \textbf{0.5347} & \textbf{0.5894} & \textbf{0.5247} 
		& \textbf{0.6265} & 0.5509 & \textbf{0.6011} & \textbf{0.5320} 
		&\textbf{0.6318} & 0.5570 & \textbf{0.6166} & \textbf{0.5447} \\
         
		\bottomrule
	\end{tabular}
	
\end{table*}

\begin{table*}[!t]
        \caption{Results under the five-shot setting using Criminal as the source data. Underlined values denote the optimal results of baselines.}
	\label{exp:fiveshot}
	\centering
	\begin{tabular}{lcccc|cccc|cccc} \toprule
		Datasets  
		& \multicolumn{4}{c}{\textbf{Criminal-S} $\rightarrow$ \textbf{NCCP}}
		& \multicolumn{4}{c}{\textbf{Criminal-M} $\rightarrow$ \textbf{NCCP}} 
		& \multicolumn{4}{c}{\textbf{Criminal-L} $\rightarrow$ \textbf{NCCP}} \\
		\cmidrule(lr){2-5} \cmidrule(lr){6-9} \cmidrule(lr){10-13}
		Metrics  & Acc.  & MP    & MR    & F1    & Acc.  & MP    & MR    & F1    & Acc.  & MP    & MR    & F1 \\
		\midrule 
            FewShot + T
		& 0.5815 & 0.4733 & 0.5219 & 0.4488 
		& 0.6188 & 0.5024 & 0.5409 & 0.4813 
		& 0.6206 & \underline{0.5302} & 0.5634 & 0.5034 \\
            CECP + T
		& 0.5662 & 0.4659 & 0.4858 & 0.4384 
		& 0.5943 & 0.4648 & 0.4846 & 0.4451 
		& 0.5920 & 0.4918 & 0.5162 & 0.4635 \\
            \midrule
		  FADA  
		& 0.5946 & 0.4933 & 0.5820 & 0.4856 
		& 0.5937 & 0.5044 & 0.5966 & 0.4981 
		& 0.6055 & 0.5090 & 0.5979 & 0.5014 \\
            $d$-SNE  
		& 0.5910 & 0.4898 & 0.5863 & 0.4847 
		& 0.5888 & 0.4779 & 0.5756 & 0.4769 
		& 0.6048 & 0.4902 & 0.6014 & 0.4931 \\
            DAGE  
		& \underline{\textbf{0.6268}} & \underline{\textbf{0.5134}} & \underline{0.6028} & \underline{0.5163} 
		& \underline{0.6334} & \underline{0.5103} & \underline{0.6116} & \underline{0.5175} 
		& \underline{0.6371} & 0.5129 & \underline{0.6023} & \underline{0.5161} \\
		\midrule
		\textbf{DLCCP+GRU}  
		& 0.6244 & 0.5113 & \textbf{0.6065} & \textbf{0.5195} 
		& \textbf{0.6349} & \textbf{0.5230} & \textbf{0.6171} & \textbf{0.5308} 
		& \textbf{0.6398} & \textbf{0.5421} & \textbf{0.6280} & \textbf{0.5489} \\
            \midrule
            BERTL+T
		& 0.6228 & 0.5119 & 0.5599 & 0.4995 
		& 0.6234 & 0.5149 & 0.5432 & 0.5020 
		& 0.6500 & 0.5627 & 0.5643 & 0.5314 \\
		  LeapBank+T
		& \underline{0.6284} & \underline{0.5519} & \underline{0.5939} & \underline{0.5324} 
		& \underline{0.6533} & \underline{0.5457} & \underline{0.5877} & \underline{0.5338} 
		& \underline{0.6531} & \underline{\textbf{0.5763}} & \underline{0.5950} & \underline{0.5413} \\
            \midrule
		  \textbf{DLCCP+BERT}
		& \textbf{0.6553} & \textbf{0.5552} & \textbf{0.6448} & \textbf{0.5582} 
		& \textbf{0.6569} & \textbf{0.5624} & \textbf{0.6521} & \textbf{0.5670} 
		& \textbf{0.6542} & 0.5625 & \textbf{0.6622} & \textbf{0.5704} \\
		\bottomrule
	\end{tabular}
	
\end{table*}

Following the work of \cite{cecp,double_layer}, we employ Accuracy (Acc.), Macro-Precision (MP), Macro-Recall (MR), and Macro-F1 (F1) as the evaluation metrics. Like Criminal, NCCP is also an imbalanced dataset. Hence, Acc. might be dominated by high-frequency charges, while MP, MR, and F1 are fairer. We train and test each model 5 times, and finally report the average results. 

\subsection{Results}

\begin{table*}[!t]
        \caption{Results under the one-shot setting using CAIL as the source data. Underlined values denote the optimal results of baselines.}
	\label{exp:oneshot cail}
	\centering
	\begin{tabular}{lcccc|cccc} \toprule
		Datasets  
		& \multicolumn{4}{c}{\textbf{CAIL-Small} $\rightarrow$ \textbf{NCCP}}
		& \multicolumn{4}{c}{\textbf{CAIL-Big} $\rightarrow$ \textbf{NCCP}}  \\
		\cmidrule(lr){2-5} \cmidrule(lr){6-9} 
		Metrics  & Acc.  & MP    & MR    & F1    & Acc.  & MP    & MR    & F1   \\
		\midrule
            FewShot
            & 0.5527 & 0.5883 & 0.5677 & 0.5056 
		& 0.6265 & 0.6274 & 0.5612 & 0.5447 \\
		CECP  
		& 0.5170 & 0.5854 & 0.5461 & 0.4962 
		& 0.6190 & 0.5880 & 0.5778 & 0.5500 \\
            FewShot+T
		& 0.5711 & 0.5595 & 0.5789 & 0.5085 
		& 0.6436 & \underline{\textbf{0.6413}} & 0.5784 & 0.5605 \\
            CECP+T
		& 0.5469 & 0.5646 & 0.5768 & 0.5077 
		& 0.6302 & 0.6315 & 0.5756 & 0.5660 \\
            \midrule
		  FADA  
		& \underline{0.6230} & 0.5584 & \underline{0.6464} & 0.5482 
		& \underline{\textbf{0.6587}} & 0.5994 & \underline{0.6374} & \underline{0.5752} \\
            $d$-SNE  
		& 0.6185 & 0.5778 & 0.6214 & \underline{0.5528} 
		& 0.5305 & 0.4659 & 0.4967 & 0.4344 \\
            DAGE  
		& 0.5959 & \underline{\textbf{0.5863}} & 0.6145 & 0.5473 
		& 0.6358 & 0.5454 & 0.5984 & 0.5296 \\
		\midrule
		\textbf{DLCCP+GRU}  
		& \textbf{0.6390} & 0.5810 & \textbf{0.6596} & \textbf{0.5619} 
		& 0.6484 & 0.5775 & \textbf{0.6628} & \textbf{0.5798} \\

            \midrule
            BERTL  
		& 0.5070 & 0.5804 & 0.5060 & 0.4699 
		& 0.6287 & 0.5679 & 0.5355 & 0.5118 \\  
		  LeapBank
		& 0.5457 & \underline{\textbf{0.6098}} & 0.5788 & 0.5247 
		& 0.6320 & 0.6130 & 0.5420 & 0.5364 \\  
            RERTL+T
		& 0.5538 & 0.5364 & 0.5698 & 0.4957 
		& 0.5943 & 0.6387 & 0.5249 & 0.5223 \\  
		  LeapBank+T
		& \underline{0.5657} & 0.5937 & \underline{0.5936} & \underline{0.5309} 
		& \underline{0.6321} & \underline{0.6403} & \underline{0.5805} & \underline{0.5590} \\
            \midrule
            \textbf{DLCCP+BERT}  
		& \textbf{0.5912} & 0.5648 & \textbf{0.6207} & \textbf{0.5378} 
		& \textbf{0.6457} & \textbf{0.5950} & \textbf{0.6634} & \textbf{0.5824} \\

		\bottomrule
	\end{tabular}
	
\end{table*}

\begin{table*}[!h]
        \caption{Results under the five-shot setting using CAIL as the source data. Underlined values denote the optimal results of baselines.}
	\label{exp:fiveshot cail}
	\centering
	\begin{tabular}{lcccc|cccc} \toprule
		Datasets  
		& \multicolumn{4}{c}{\textbf{CAIL-Small} $\rightarrow$ \textbf{NCCP}}
		& \multicolumn{4}{c}{\textbf{CAIL-Big} $\rightarrow$ \textbf{NCCP}} \\
		\cmidrule(lr){2-5} \cmidrule(lr){6-9} 
		Metrics  & Acc.  & MP    & MR    & F1    & Acc.  & MP    & MR    & F1 \\
		\midrule 
            FewShot + T
		& 0.6039 & 0.5678 & 0.6316 & 0.5390 
		& 0.6554 & \underline{0.6068} & 0.5887 & 0.5541 \\  
            CECP + T
		& 0.5916 & 0.5759 & 0.6140 & 0.5410 
		& 0.6475 & 0.5987 & 0.6160 & 0.5157 \\
            \midrule
		  FADA  
		& 0.6529 & 0.5790 & 0.6739 & 0.5757 
		& \underline{0.6623} & 0.5699 & \underline{0.6591} & \underline{0.5661} \\
            $d$-SNE  
		& \underline{0.6765} & \underline{\textbf{0.6151}} & \underline{\textbf{0.6997}} & \underline{\textbf{0.6157}} 
		& 0.6414 & 0.5200 & 0.6360 & 0.5318 \\  
            DAGE  
		& 0.6669 & 0.5937 & 0.6857 & 0.5982 
		& 0.6605 & 0.5494 & 0.6195 & 0.5437 \\

		\midrule
		\textbf{DLCCP+GRU}  
		& \textbf{0.6821} & 0.5888 & 0.6938 & 0.6011 
		& \textbf{0.6881} & \textbf{0.6151} & \textbf{0.6857} & \textbf{0.6165} \\
            \midrule
            BERTL+T
		& 0.5942 & 0.5568 & 0.6036 & 0.5283 
		& \underline{0.6673} & 0.6040 & \underline{0.6052} & \underline{0.5738} \\
		  LeapBank+T
		& \underline{0.6030} & \textbf{\underline{0.5856}} & \underline{0.6319} & \underline{0.5529} 
		& 0.6449 & \textbf{\underline{0.6458}} & 0.5763 & 0.5618 \\
		  \textbf{DLCCP+BERT}
		& \textbf{0.6466} & 0.5847 & \textbf{0.6830} & \textbf{0.5795} 
		& \textbf{0.6715} & 0.5982 & \textbf{0.6855} & \textbf{0.6004} \\
  
		\bottomrule
	\end{tabular}
	
\end{table*}

\subsubsection{Criminal}

When employing the Criminal dataset as the source data, Table \ref{exp:oneshot} and Table \ref{exp:fiveshot} respectively show the experimental results on NCCP under the one-shot setting (one sample per class from NCCP for training) and the five-shot setting (five samples per class from NCCP for training). The RNN-based methods (top portion of the table) and the transformer-based methods (bottom portion of the table) are compared separately. We can obtain the following observations: 
(1) FSDA methods (FADA, $d$-SNE, DAGE, and DLCCP) that can alleviate the domain discrepancy usually perform better than charge prediction models designed for the source domain (FewShot, CECP, BERTL, and LeapBank). However, these traditional FSDA methods do not match up to the performance of the proposed DLCCP across a wide range of experimental settings and evaluation metrics. This discrepancy arises from their failure to consider the negative impact of text style information contained in latent features and their inability to extract and align element-level information across domains. This demonstrates a specialized FSDA model is essential for charge prediction with non-PLLS data;  
(2) in the one-shot setting, FADA shows better performance over baselines. The reason might be that the adversarial strategy can better confuse different domains. DAGE shows better performance in the five-shot setting. This improvement may be attributed to its utilization of graph embedding, which can capture the structure of the target data better when the amount of target data increases. Yet our model performs well in both settings; 
(3) when the target data is included in the training process, the performance of traditional methods designed for charge prediction can be somewhat enhanced. This is primarily attributed to two factors. On one hand, the source data exhibits a significant imbalance in class distribution, which is mitigated by the incorporation of target data. On the other hand, these models may also capture, to a certain extent, distributional characteristics of the target domain data during training. Nevertheless, these methods still encounter challenges in addressing the discrepancy between the two domains, thereby resulting in their performance being inferior to the FSDA method;
(4) the transformer-based methods outperform RNN-based methods. This shows that the cross-domain charge prediction can also benefit from pre-trained models; 
(5) we use t-test with significance level 0.05 to test the significance of performance difference. Results show that in most cases DLCCP significantly outperforms the baselines on both settings.

\subsubsection{CAIL}

Table \ref{exp:oneshot cail} and Table \ref{exp:fiveshot cail} respectively show the experimental results on NCCP under the one-shot setting and the five-shot setting when the CAIL dataset is employed as the source data. We have observed similar experimental findings to those obtained when using the Criminal dataset as the source data. For instance, FSDA methods generally outperform traditional charge prediction methods, and incorporating target data into the training process leads to certain improvements for traditional charge prediction methods. Furthermore, we have uncovered additional novel findings and drawn new conclusions: 
(1) with CAIL-Small as the source data, transformer-based methods do not exhibit performance improvements and even underperform compared to RNN-based methods. This could be attributed to the limited scale of the source data, leading to the over-fitting problem of the transformer-based models;
(2) as the amount of source data increases, the performance of $d$-sne exhibits significant degradation. This is likely due to the inherent characteristics of $d$-sne. It attempts to minimize (maximize) the largest (smallest) distance between samples of the same (different) class from two domains, which creates a sensitivity to outliers. This sensitivity arises because the method's emphasis on these extreme distances may lead it to inadvertently accentuate outliers. To validate this, we employ the Isolation Forest \cite{isolation} to detect outliers in CAIL-Small and CAIL-Big datasets, resulting in outlier proportions of 0.208\% and 0.251\%, respectively. Furthermore, the FADA and DAGE also exhibit this phenomenon with slightly milder declines. In contrast, the proposed DLCCP demonstrates robustness.

\subsection{Analysis}

We further conduct a variety of experiments to analyze the performance of DLCCP from different perspectives. In the following experiments, we employ the Criminal-S as the source data.

\begin{figure*}[!t]
	\centering
	\includegraphics[width=0.96\textwidth]{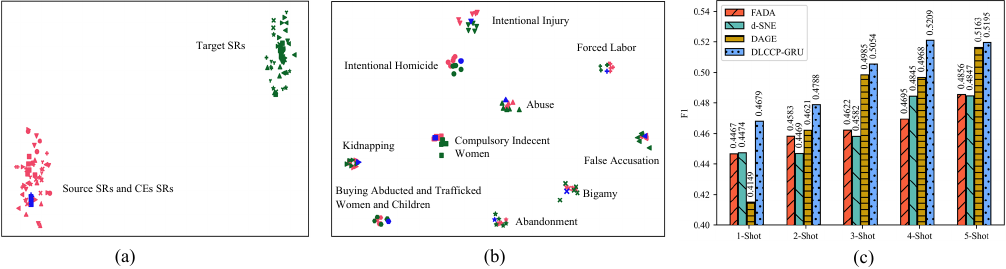}
	\caption{(a) t-SNE plot of SRs; (b) t-SNE plot of instance-level CRs. In (a) and (b), different shapes of markers indicate different charges, while the red, green, and blue indicate the source domain, target domain, and CEs, respectively; (c) The F1 values with different shot numbers.}
	\label{exp:vis}
\end{figure*}

\begin{figure*}[!h]
	\centering
	\includegraphics[width=0.98\textwidth]{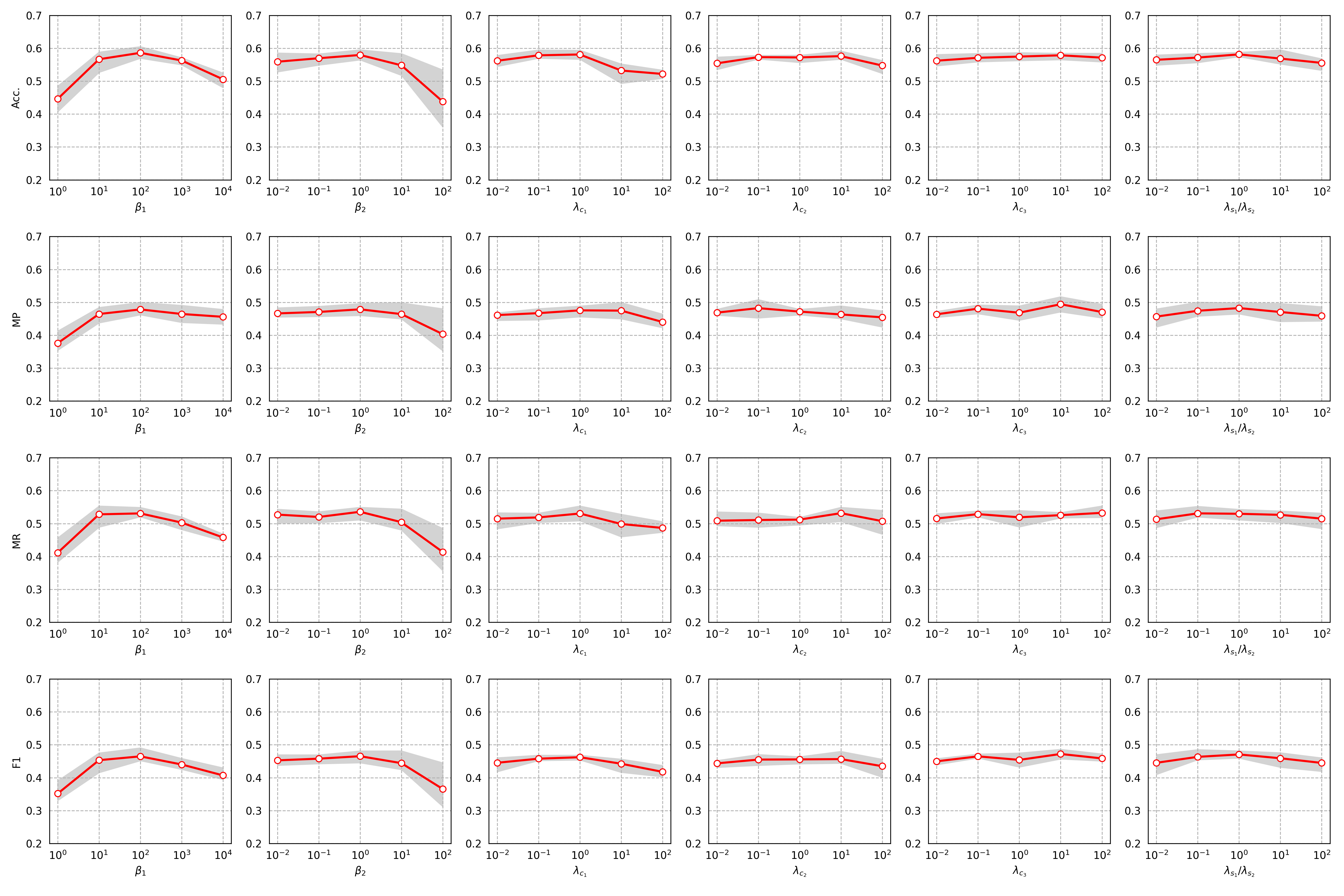}
	\caption{Comparison of different parameters tested under the one-shot setting. Each individual subgraph shows the experimental results when a specific parameter takes different values while keeping other parameters fixed.}
	\label{exp:para}
\end{figure*}

\subsubsection{Visualization of the Latent Space} To understand how CRs and SRs are disentangled, we employ the t-SNE to visualize their distributions in the latent space. Here we select 10 confusing charges which are all related to violating human rights. The t-SNE plots of SRs and CRs are shown in Fig. \ref{exp:vis}(a) and Fig. \ref{exp:vis}(b) respectively. We can see that: (1) the SRs with the same text style are clustered and the SRs with different text styles are separated distinctly; (2) CRs with the same class from different domains are noticeably clustered around the CR of CEs with corresponding charge. These observations demonstrate that our approach can disentangle CRs and SRs effectively.

\subsubsection{Influence of the Number of Shots} Moreover, we evaluate how the model performance is impacted by the number of samples per class (denoted as $n$) from the target domain.  Here we employ the F1 score as the metric. The results of FSDA methods are shown in Figure \ref{exp:vis}(c). It is evident that the F1 values of most of the methods increase as $n$ increases. And DLCCP produces the best results in all scenarios.

\begin{table}[!t]
        \caption{Results of ablation studies.}
	\label{exp:ablation}
	\centering
	\begin{tabular}{lcccc} \toprule
		Metrics & Acc. & MP    & MR & F1 \\
		\midrule
            \textbf{DLCCP+GRU} & \textbf{0.5795} & \textbf{0.4775} & \textbf{0.5342} & \textbf{0.4679}    \\
            \midrule
		-DCS  &  0.5771 & 0.4681 & 0.5291 & 0.4610  \\    
            -ELA  &  0.5736 & 0.4698 & 0.5229 & 0.4589  \\ 
		\bottomrule
	\end{tabular}
	
\end{table}

\subsubsection{Parameters and Performance Stability Analysis}  

In this part, we evaluate the performance and stability of DLCCP across different settings of hyperparameters. These experiments are conducted under the one-shot setting, and we train and test our model five times. While varying one parameter, we keep the other parameters fixed. Particularly, we set $s_1=s_2$. The corresponding experimental results are presented in Fig. \ref{exp:para}. The variation magnitudes of the results are illustrated by the shaded regions in the figure. It can be observed that, under appropriate hyperparameters, our model not only achieves commendable predictive performance but also exhibits notable stability.

\subsubsection{Ablation Study} We conduct ablation experiments to demonstrate the effectiveness of two key mechanisms within the DLCCP framework: (1) the disentanglement of CRs and SRs (DCS). To explore the impact of this, we intentionally disregard all constraints concerning SRs to form a variant denoted as ``-DCS'' (2) the element-level alignment (ELA). In this regard, we only perform the instance-level alignment ($\lambda_{c_1}=0$) and form a variant denoted as ``-ELA''. The experiments are operated under the one-shot setting and the results are presented in Table \ref{exp:ablation}. Notably, we can observe an apparent decrease in performance for both the "-DCS" and "-ELA" variants. This observation confirms the pivotal role and effectiveness of the DCS and ELA mechanisms.

\section{Conclusion}
\noindent In this paper, we propose a novel FSDA method, Disentangled Legal Content for Charge Prediction (DLCCP), tailored for charge prediction on the non-PLLS data. It disentangles the fact descriptions (CEs) into CRs and SRs, thereby mitigating the domain discrepancy between the PLLS data and non-PLLS data. Furthermore, it aligns element-level CRs, which are essential to our task, and ultimately predicts charges based on the GRU-derived instance-level CRs. We carefully design different optimization goals for the content space and the style space of fact descriptions and CEs, facilitating the learning of better domain-invariant and legal-specific CRs. Our extensive experiments substantiate the superior effectiveness of DLCCP in comparison to competitive baselines.

\section*{Acknowledgments}
\noindent This research was supported by the National Natural Science Foundation of China (Grant Nos.62133012, 61936006, 62073255), the Key Research and Development Program of Shaanxi (Program No.2020ZDLGY04-07), and Innovation Capability Support Program of Shaanxi (Program No. 2021TD-05).

\bibliographystyle{IEEEtran}
\bibliography{dlccp}

\begin{IEEEbiography}[{\includegraphics[width=0.9in, height=1.25in,clip,keepaspectratio]{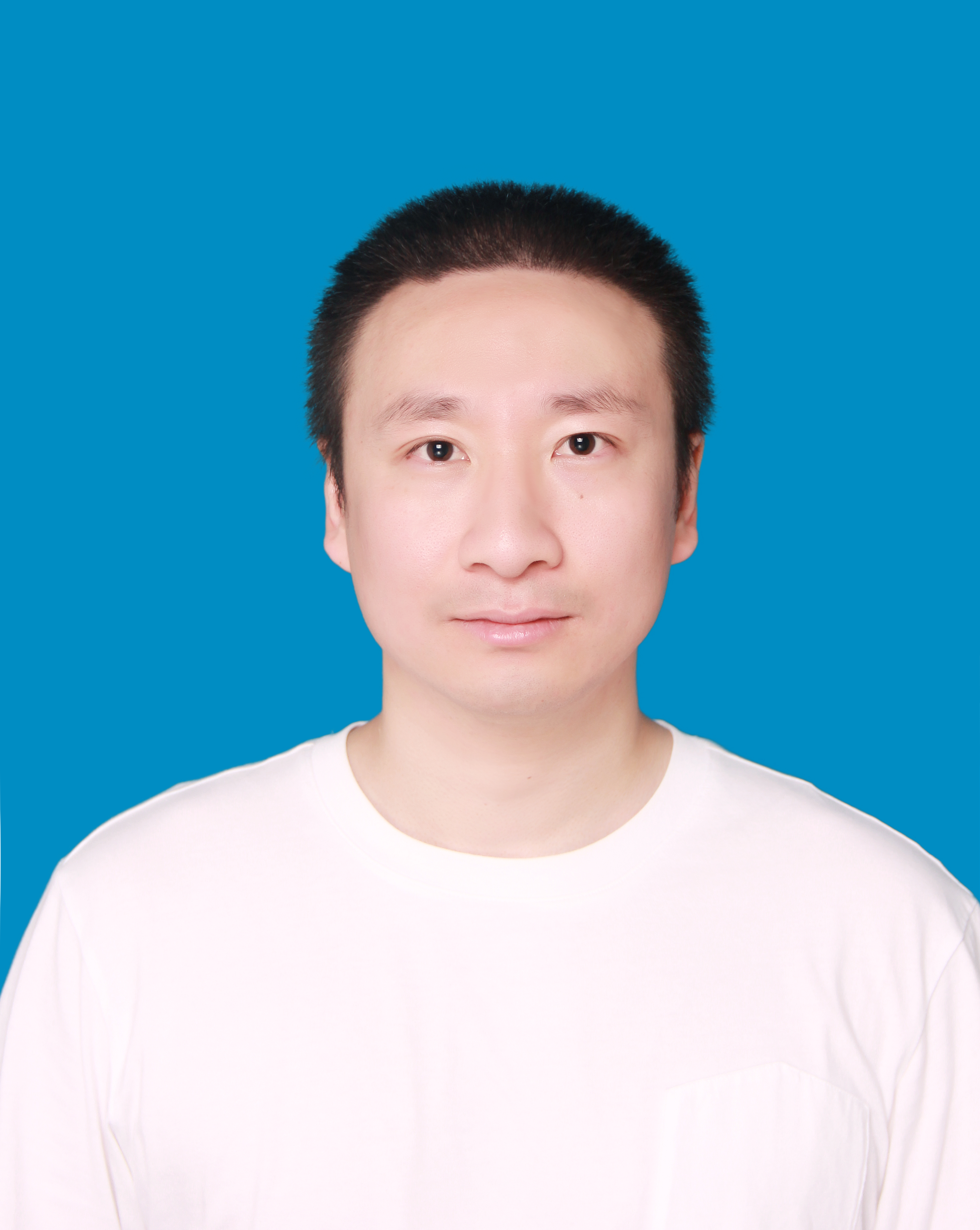}}]{Jie Zhao} received the B.S. degree from Southwest University in 2016 and the M.S. degree from Xidian University in 2019. He is now working toward the PhD degree in the School of Computer Science and Technology, Xidian University. His research interests include natural language processing and machine learning.
\end{IEEEbiography}

\begin{IEEEbiography}[{\includegraphics[width=0.9in, height=1.25in,clip,keepaspectratio]{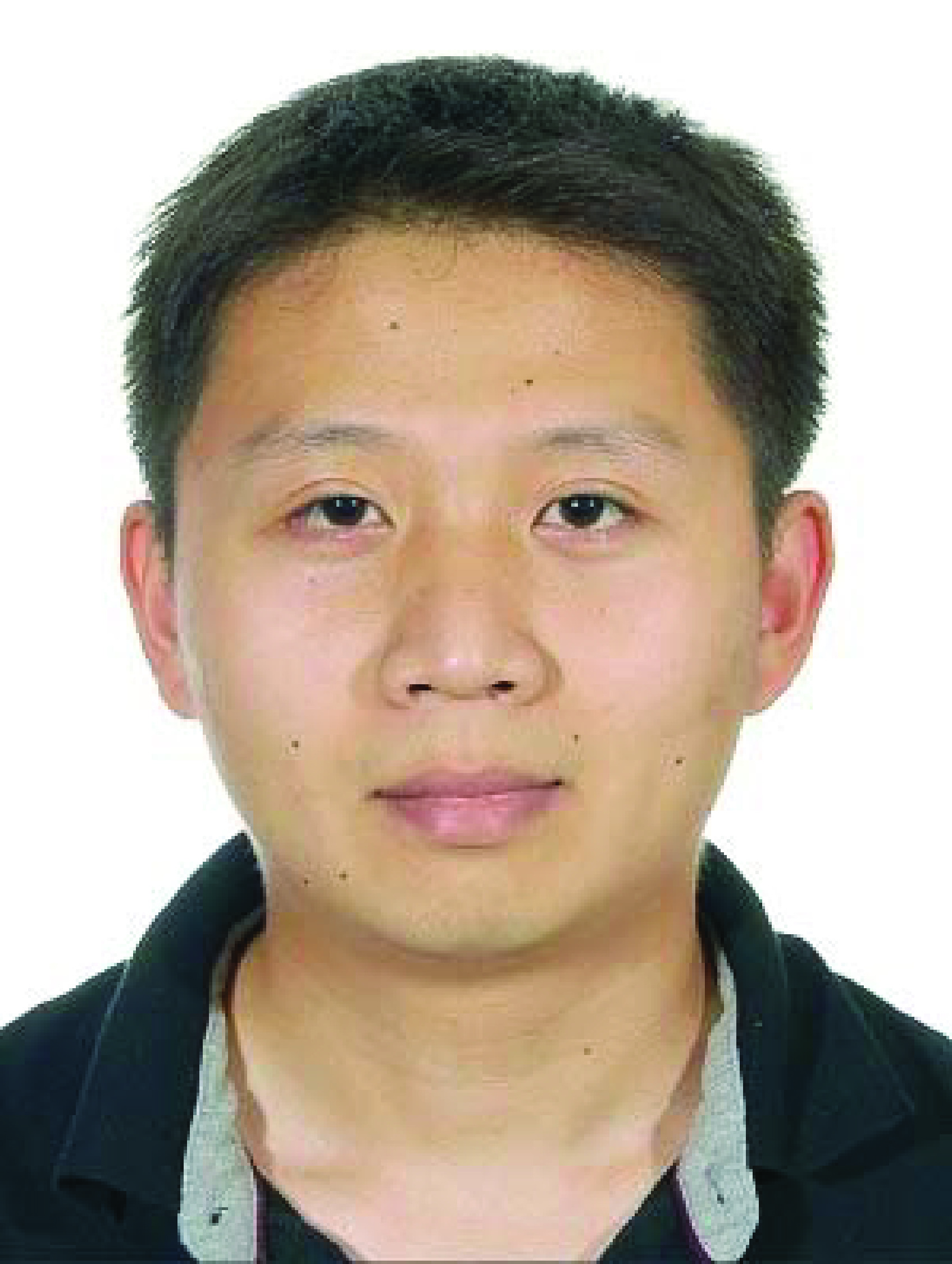}}]{Ziyu Guan} received the B.S. and Ph.D. degrees in Computer Science from Zhejiang University, Hangzhou China, in 2004 and 2010, respectively. He had worked as a research scientist in the University of California at Santa Barbara from 2010 to 2012, and as a professor in the School of Information and Technology of Northwest University, China from 2012 to 2018. He is currently a professor with the School of Computer Science and Technology, Xidian University. His research interests include attributed graph mining and search, machine learning, expertise modeling and retrieval, and recommender systems.
\end{IEEEbiography}

\begin{IEEEbiography}[{\includegraphics[width=0.9in, height=1.25in,clip,keepaspectratio]{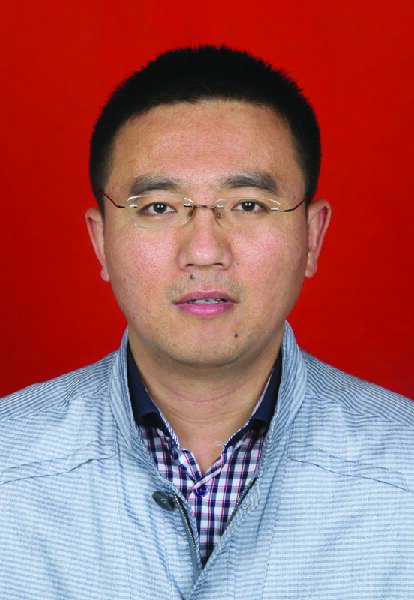}}]{Wei Zhao} received the B.S., M.S. and Ph.D. degrees from Xidian University, Xi’an, China, in 2002, 2005 and 2015, respectively. He is currently a professor in the School of Computer Science and Technology at Xidian University. His research direction is pattern recognition and intelligent systems, with specific interests in attributed graph mining and search, machine learning, signal processing and precision guiding technology.
\end{IEEEbiography}

\begin{IEEEbiography}[{\includegraphics[width=0.9in, height=1.25in,clip,keepaspectratio]{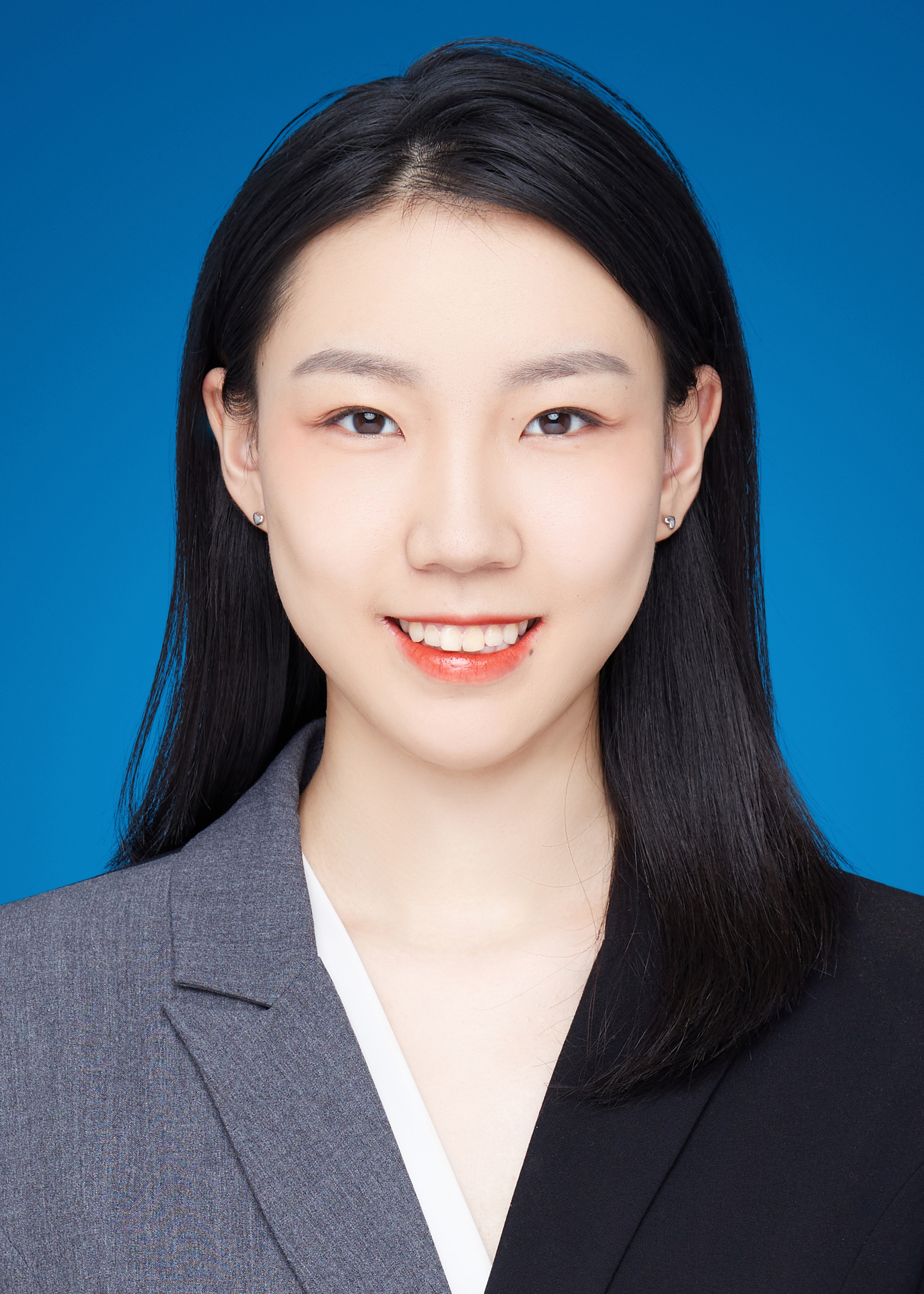}}]{Yue Jiang} received the B.S. degree from Hefei University  of Technology in 2022. She is now working toward the M.S. degree in the School of Computer Science and Technology, Xidian University. Natural language processing is one of her research interests.
\end{IEEEbiography}

\begin{IEEEbiography}[{\includegraphics[width=0.9in, height=1.25in,clip,keepaspectratio]{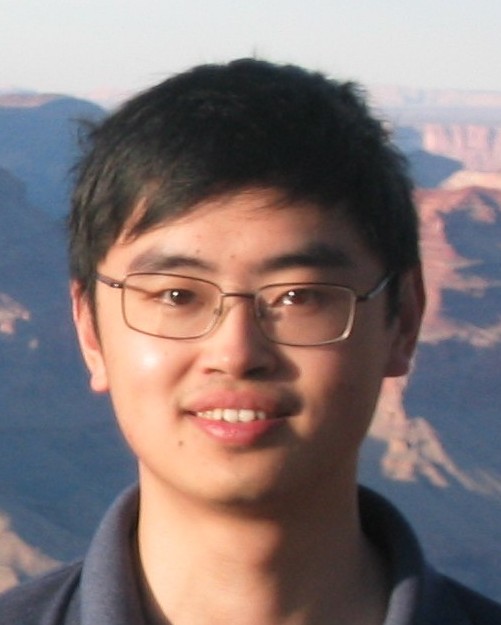}}]{Xiaofei He} received the BS degree in computer science from Zhejiang University, China, in 2000 and the PhD degree in computer science from the University of Chicago in 2005. He was a research scientist with Yahoo! Research Labs, Burbank, CA. He is currently a professor with the State Key Lab of CAD\&CG, Zhejiang University, China. His research interests include machine learning, information retrieval, and computer vision.
\end{IEEEbiography}

\vfill

\end{document}